\documentclass[11pt]{article}
\usepackage{geometry}
\usepackage{coling2020}
\usepackage{times}
\usepackage{url}
\usepackage{latexsym}
\usepackage{microtype}
\usepackage[utf8x]{inputenc}
\usepackage{url}
\usepackage{times}
\usepackage{latexsym}
\usepackage{todonotes}
\usepackage{amsthm}
\usepackage{adjustbox}
\usepackage{multirow}
\usepackage{multicol}
\usepackage{arydshln}
\usepackage{float}
\usepackage{graphicx}
\usepackage{subcaption}
\usepackage[skip=5pt]{caption}
\usepackage{mathtools}
\usepackage{amsmath}
\usepackage{amssymb}
\usepackage{enumerate}

\colingfinalcopy

\title{IMS at SemEval-2020 Task 1:\\
How low can you go? Dimensionality in Lexical Semantic Change Detection
}

\author{Jens Kaiser, Dominik Schlechtweg\thanks{~~As an organizer of SemEval-2020 Task 1 DS contributed only after the evaluation phase.}, Sean Papay, Sabine Schulte im Walde \\
Institute for Natural Language Processing, University of Stuttgart\\
\small
{\tt \{jens.kaiser,schlecdk,sean.papay,schulte\}@ims.uni-stuttgart.de}
}
\date{}

\begin{document}
\maketitle
\begin{abstract}
    We present the results of our system for SemEval-2020 Task 1 that exploits a commonly used lexical semantic change detection model based on Skip-Gram with Negative Sampling. Our system focuses on Vector Initialization (VI) alignment, compares VI to the currently top-ranking models for Subtask 2 and demonstrates that these can be outperformed if we optimize VI dimensionality. We demonstrate that differences in performance can largely be attributed to model-specific sources of noise, and we reveal a strong relationship between dimensionality and frequency-induced noise in VI alignment. Our results suggest that lexical semantic change models integrating vector space alignment should pay more attention to the role of the dimensionality parameter.
\end{abstract}

\section{Introduction}
\blfootnote{This work is licensed under a Creative Commons Attribution 4.0 International Licence. Licence details:\\ \url{http://creativecommons.org/licenses/by/4.0/.}}
Lexical Semantic Change (LSC) Detection has drawn increasing attention in recent years \cite{2018arXiv181106278T,kutuzov-etal-2018-diachronic}. SemEval-2020 Task 1 provides a multi-lingual evaluation framework to compare the variety of proposed model architectures \cite{schlechtweg2020semeval}. An important component of high-performance LSC detection models is an alignment method to make semantic vector spaces comparable across time. In this paper we focus on a particular alignment method for type embeddings, Vector Initialization (VI), and how its performance interacts with vector dimensionality. We compare VI to two further state-of-the-art alignment methods, Orthogonal Procrustes (OP) and Word Injection (WI), which have shown high performance in previous studies \cite{Hamilton2016b,Schlechtwegetal19,dubossarskyetal19} and are also used in the top-ranking systems for Subtask 2. A systematic comparison of performance across dimensionalities $d$ reveals that the optimal $d$ of the models on the SemEval test data is lower than in standard choices, and that VI's performance strongly depends on $d$, showing large drops for high dimensionalities. We demonstrate that this effect is correlated with the amount of frequency noise picked up by VI, i.e., the degree to which cosine distances between vectors reflect frequency differences between words rather than semantic differences. If properly tuned regarding dimensionalities and noise, VI outperforms OP and WI as alignment method.

\section{Related Work}

The semantic representations we test fall into the large body of work on distributional semantic vector space models \cite{Turney:2010} and represent specific instances of type-based word embeddings \cite{Mikolov13a}. The need for vector space alignment in LSC detection is shared with bilingual lexicon induction \cite{Ruder2019} and term extraction \cite{Haettyetal20} where corpus-specific semantic representations need to be mapped to common coordinate axes.

Alignment techniques introduce varying levels of noise \cite{dubossarskyetal19}, and the noise level (the signal-to-noise-ratio) determines the optimal dimensionality of word embeddings \cite{Yin2018}. We regard information contained within the semantic representation capturing anything but semantic relations between words as noise (e.g. word frequency). Sources of noise include the corpora, the representation method and alignment techniques. Consequently, a specific semantic representation learning algorithm (such as Skip-Gram with Negative Sampling) may have a different optimal dimensionality depending on the alignment technique it relies on. Up to now, previous research on LSC detection has not paid much attention to this relationship between dimensionality and noise: models with different susceptibilities to noise have typically been tested without varying the dimensionality \cite{Hamilton2016b,dubossarskyetal19,Schlechtwegetal19,Shoemark2019}.

\section{System overview}

Most models in LSC detection combine three sub-systems: (i) creating semantic word representations, (ii) aligning them across corpora, and (iii) measuring differences between the aligned representations \cite{Schlechtwegetal19}. Semantic representations can either be token-based, keeping one representation (e.g. a vector) per word use \cite[e.g.]{Hu19}, or type-based, collapsing information from different uses into one representation \cite[e.g.]{Hamilton2016b}. The alignment step is needed mostly for vector space models, which might otherwise introduce arbitrary orthogonal transformations to the vector spaces they produce \cite{Hamilton2016b}.

Our system focuses on the type-based Skip-gram Negative Sampling (SGNS) with Vector Initialisation alignment (VI) and Cosine Distance (CD). We chose this method due to its surprisingly good performance with $d=5$ in a student shared-task project \cite{Ahmad2020}\footnote{These referenced results were achieved by us after experimenting with extreme and unusual parameter choices.} and compare it to two variations of the alignment method, (i)~Orthogonal Procrustes (OP) and (ii)~Word Injection (WI). For our experiments we use the code provided by \newcite{Schlechtwegetal19}.\footnote{\url{https://github.com/Garrafao/LSCDetection}}

\subsection{Semantic Representation}

SGNS is a shallow neural network trained on pairs of word co-occurrences extracted from a corpus with a symmetric window. It represents each word $w$ and each context $c$ as a $d$-dimensional vector to solve
\begin{equation}
\arg\max_\theta \sum_{(w,c)\in D} \log \sigma(v_c \cdot v_w) + \sum_{(w,c) \in D'} \log \sigma (-v_c \cdot v_w),
\end{equation}
where $\sigma(x) = \frac{1}{1+e^{-x}}$, $D$ is the set of all observed word-context pairs and $D'$ is the set of randomly generated negative samples \cite{Mikolov13a,Mikolov13b,GoldbergL14}. The optimized parameters $\theta$ are $v_{w_i}$ and $v_{c_i}$ for $i\in 1,...,d$. $D'$ is obtained by drawing $k$ contexts from the empirical unigram distribution $P(c) = \frac{\#(c)}{|D|}$ for each observation of $(w,c)$, cf. \newcite{Levy2015}. After training, each word $w$ is represented by its word vector $v_{w}$. To keep our results comparable to previous research \cite{Hamilton2016b,Schlechtwegetal19} we chose common settings for most of the hyper-parameters. We decided on a symmetrical context window of size $10$, initial learning rate $\alpha$ of $0.025$, number of negative samples $k=5$ and no sub-sampling. Depending on corpus size we trained the model for either 5 (German, Swedish) or 30 epochs $e$ (English, Latin).\footnote{We tried alternative numbers of epochs with mixed results.} As we focus on the effect of dimensionality, each experiment was performed for each $d \in \{5, 10, 25, 50, 80, 150, 200, 250, 300, 350, 500, 750, 1000\}$. Prior to the shared task application we validated all models with these hyper-parameters on the German DURel dataset \cite{Schlechtwegetal18}.

\subsection{Alignment}
\label{sec:alignment}

\paragraph{Vector Initialisation.} In VI we first train the SGNS model on one corpus and then use these vectors to initialize the vectors for training on the second corpus \cite{Kim14}. The motivation of this procedure is that if a word is used in similar contexts in both corpora, the second training step will not change the initial word vector much, while more different contexts will lead to a greater change of the vector. SGNS represents each word by two vectors, a word vector and a context vector. The former is modified when a word occurs as target $w$ in a target-context pair $(w,c)$, while the latter is modified when it occurs as context $c$. While \newcite{Schlechtwegetal19} only initialize the word vectors on the first model and context vectors randomly, we also initialize context vectors on the first model, as done by \newcite{Ahmad2020}. In this way, we expect to introduce considerably less noise to the vectors in the second corpus.

\paragraph{Orthogonal Procrustes.} SGNS is trained on each corpus separately, resulting in matrices $A$ and $B$. To align them we follow \newcite{Hamilton2016b} and calculate an orthogonally-constrained matrix $W^*$: 
\begin{equation}
W^* =\underset{W \in O(d)}{\arg\min} \left\lVert B W - A\right\lVert_F
\end{equation}
where the $i$-th row in matrices $A$ and $B$ correspond to the same word. Using $W^*$ we get the aligned matrices $A^{OP} = A$ and $B^{OP} = BW^*$. Prior to this alignment step we length-normalize and mean-center both matrices \cite{Artetxe2017,Schlechtwegetal19}.

\paragraph{Word Injection.} The sentences of both corpora are shuffled into one joint corpus, but all occurrences of target words are substituted by the target word concatenated with a tag indicating the corpus it originated from \cite{ferrari2017detecting,Schlechtwegetal19}. This leads to the creation of two vectors for each target word in one vector space, while non-target words receive only one vector encoding information from both corpora. This is very similar to Temporal Referencing (TR) \cite{dubossarskyetal19}, the difference being that with TR, the target-context pairs used for training never contain tagged target words as contexts but rather the genuine (untagged) words.

\subsection{Measures}

To quantify semantic change on the aligned vector representations, we use two vector similarity measures. For Subtask 1 we apply Local Neighborhood Distance (LND), as it showed superior performance to CD for binary change detection in \newcite{SchlechtwegWalde20}. LND is based on second-order cosine similarity and measures to which extent $\vec{x}$ and $\vec{y}$'s distances to a union of their $k$ nearest neighbors differ \cite{Hamilton2016a}. Similar to \newcite{Hamilton2016a} we chose $k=25$. We split the LND scores into two equally sized groups; the group containing the high values was labelled as 1 (change). For Subtask 2 we use a simple cosine distance \cite{SaltonMcGill1983}.

\section{Experimental setup}

SemEval-2020 Task 1 comprises a binary classification task (Subtask 1) and a ranking task (Subtask 2) on data from four languages: English, German, Latin and Swedish \cite{schlechtweg2020semeval}. Subtask 1 asks participants to decide which target words lost or gained senses between corpora from two time periods $t_1$ and $t_2$, and which ones did not. Subtask 2 asks participants to rank a set of target words according to their degree of LSC (change in sense frequency distribution) between $t_1$ and $t_2$. The tasks are different in that it is possible for a word to show a high degree of LSC in Subtask 2, while not gaining or losing a sense in Subtask 1 (or vice versa). For example the German word \textit{abgebrüht} is used to describe (1) the process of cooking food in water and (2) an emotionally insensitive person. Both senses are present across time periods $t_1$ and $t_2$, but sense 1 dominates period $t_1$ while sense 2 dominates $t_2$. Thus, the number of senses has not changed, but the word has undergone significant semantic change. 

The four languages have a list of 31 to 48 target words, each annotated with values for Subtasks 1 and 2. Performance of a model is measured by accuracy and Spearman's rank-order correlation coefficient. For each of the four languages two corpora (corpus$_1$, corpus$_2$) are provided by the organizers, containing sentences from different time periods. These corpora show strong differences in terms of size, time-period and genre (see Appendix \ref{sec:corpora}), posing a very heterogeneous, challenging setting for evaluation and parameter tuning.\footnote{For pre-processing details see Appendix \ref{sec:process}.}

\begin{table}
	\center
	\small
	\begin{adjustbox}{width=1.0\linewidth}
	\begin{tabular}{c| l | l | c c c c c | c c c c c }
		\hline
	  \multirow{2}{*}{\textbf{Phase}} & \multirow{2}{*}{\textbf{Team}} & \multirow{2}{*}{\textbf{Model}} &\multicolumn{5}{c |}{\textbf{Subtask 1 (accuracy)}} & \multicolumn{5}{c}{\textbf{Subtask 2 (spearman $\rho$)}} \\	
	& & & AVG & English & German & Latin & Swedish & AVG & English & German & Latin & Swedish \\\hline
	\multirow{7}{*}{\rotatebox[origin=c]{90}{\textbf{Evaluation}}} & 1. UWB & SGNS+CA & \textbf{.69} & .62 & .75 & .70 & .68 &  &  &  &  & \\
	 &2. Life-Lang. & fastText & .69 & .70 & .75 & .55 & .74 &  &  &  &  & \\
	 &3. Jia. \& Jin.  & SGNS+WI & .67 & .65 & .73 & .70 & .58 &  &  &  &  & \\\cdashline{2-13}
	 &1. UG Stud. & SGNS+OP &  &  &  &  &  & \textbf{.53} & .42 & .73 & .41 & .55\\
	 &2. Jia. \& Jin. & SGNS+WI &  &  &  &  &  & .52 & .33 & .72 & .44 & .59\\
	 &3. cs2020 & SGNS+OP &  &  &  &  &  & .50 & .38 & .70 & .40 & .54\\\cdashline{2-13}
     &13./8. IMS & \multirow{2}{*}{SGNS+VI} & \multirow{2}{*}{.60} & .54 & .69 & .55 & .61 & \multirow{2}{*}{.37} & .30 & .66 & .10 & .43\\
     &\textit{$d$ / $e$} & & & \textit{5 / 30} & \textit{5 / 30} & \textit{5 / 30} & \textit{5 / 30} & & \textit{5 / 50} & \textit{5 / 50} & \textit{5 / 50} &\textit{ 5 / 50}\\
    \hline
   \multirow{6}{*}{\rotatebox[origin=c]{90}{\textbf{Post-}}\rotatebox[origin=c]{90}{\textbf{Evaluation}}} & \multirow{5}{*}{IMS} & \multirow{2}{*}{SGNS+VI} & \multirow{2}{*}{\textbf{.73}} & .75 & .74 & .67 & .74 & \multirow{2}{*}{\textbf{.58}} & \textbf{.46} & \textbf{.78} & .39 & \textbf{.67}\\
    & \multirow{5}{*}{\textit{dim / epoch}}& & & \textit{10 / 30} & \textit{50 / 5} & \textit{80 / 30} & \textit{300 / 10}& & \textit{10 / 30} & \textit{50 / 5} & \textit{80 / 30} & \textit{300 / 10} \\
    & & \multirow{2}{*}{SGNS+OP} & \multirow{2}{*}{.65} & .62 & .80 & .50 & .68 & \multirow{2}{*}{.56} & .44 & .73 & \textbf{.41} & .64\\
    & & & & \textit{350 / 30} & \textit{300 / 5} & \textit{10 / 30} & \textit{500 / 5} & & \textit{350 / 30} & \textit{300 / 5} & \textit{10 / 30} & \textit{500 /5} \\
    & & \multirow{2}{*}{SGNS+WI} & \multirow{2}{*}{.67} & .62 & .71 & .65 & .68 & \multirow{2}{*}{.54} & .36 & .76 & .41 & .61\\
    & & & & \textit{10 / 30} & \textit{50 / 5} & \textit{10 / 30} & \textit{500 / 5} & & \textit{10 / 30} & \textit{50 / 5} & \textit{10 / 30} & \textit{500 / 5}\\
		\end{tabular}
	\end{adjustbox}
	\caption{Comparing our model to the top 3 of both subtasks in evaluation phase. Our results are annotated with dimensionality $d$ and training epochs $e$ at each entry. CA = Canonical Correlation Analysis.}
		\label{tab:scoretask2}
\end{table}

\section{Results}
Table \ref{tab:scoretask2} lists the evaluation phase scores of the top three contenders for both subtasks as well as our system. During this phase, submission scores and leaderboards were hidden. At the end of the evaluation phase the best submission (out of a maximum of ten) was put on the leaderboard for both subtasks. We only submitted results for VI with $d$/$e$ of 5/30, 3/50 and 8/30. The very low choice for $d$ is motivated by the results found in \newcite{Ahmad2020}, where VI has high performance with $d=5$. With this exceptionally low $d$ we scored 13th in Subtask 1 and 8th in Subtask 2 out of 33 teams. Our methodology for Subtask 1 has much room for improvement but we decided to point our attention towards Subtask 2 during post-evaluation. The three best teams for Subtask 2 used models based on OP and TR/WI. Their average scores are very similar and ahead of ours. The models seem to consistently get the best performances on German, then Swedish, followed by English and Latin. With this limited picture VI alignment seems to be inferior to OP and WI. But after tuning each model to its optimal $d$, performances are barely distinguishable (see Table \ref{tab:scoretask2}, Post-Evaluation). This is most prominent for Latin. We attribute the low score (0.10) during the Evaluation phase to the Latin data-set being very challenging due to its size and heterogeneity in combination with high discrepancies between the number of vector updates in the second training step (see below). Performance significantly improved after switching training order (see Appendix \ref{sec:order}).

\subsection{Analysis}
We now tested the performance of VI, OP and WI on the four languages with varying dimensionality. Training epochs were adjusted to compensate differences in corpus size: German and Swedish were trained for 5 epochs, Latin and English for 30 epochs. We realized that VI is very sensitive to training order (see Appendix \ref{sec:order}): this was very prominent with Swedish and Latin, as corpus$_1$ and corpus$_2$ show big differences in size. To get comparable results, we switched training order for these two languages, which means instead of first training on corpus$_1$ and then corpus$_2$, we train on corpus$_2$ and then corpus$_1$.

\begin{figure}[t]
\center
    \begin{subfigure}{0.45\textwidth}
        \includegraphics[width=\linewidth]{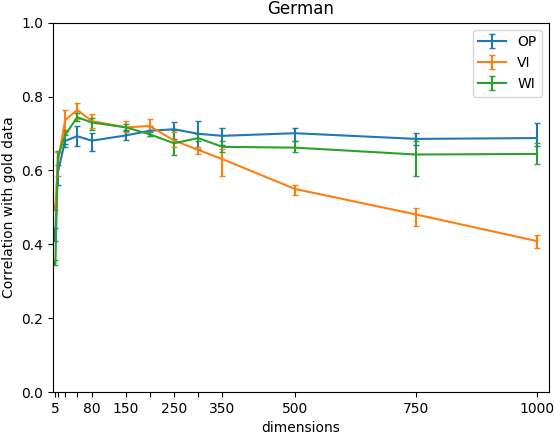}
                  \vspace{+1mm}
    \end{subfigure}    \hspace{+2mm}
    \begin{subfigure}{0.45\textwidth}
        \includegraphics[width=\linewidth]{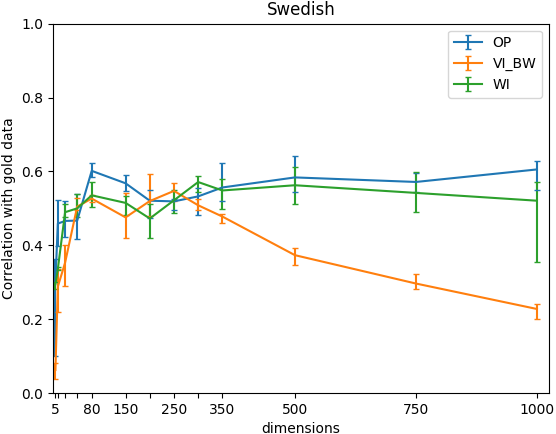}
                  \vspace{+1mm}
    \end{subfigure}
    \begin{subfigure}{0.45\textwidth}
        \includegraphics[width=\linewidth]{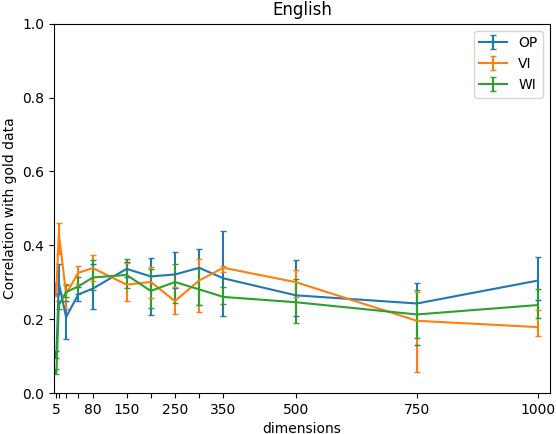}
             \end{subfigure}    \hspace{+2mm}
    \begin{subfigure}{0.45\textwidth}
        \includegraphics[width=\linewidth]{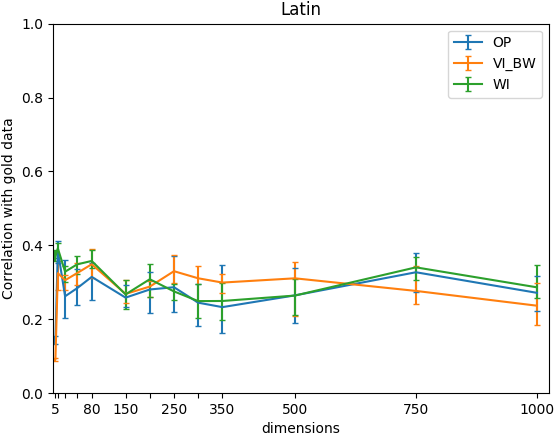}
             \end{subfigure}
    \vspace{+1mm}
    \caption{Comparing all models with varying dimensionality. The line refers to the mean of 5 runs; error bars show the max/min values.`\_BW' indicates switched training order.}
    \label{fig:task2}
\end{figure}

As shown in Figure \ref{fig:task2} all models were able to achieve high correlations for German and Swedish (0.6--0.7). On English and Latin correlation was lower (0.3--0.4),
which is probably related to corpus size, and performances are barely distinguishable, see Figure \ref{fig:task2} bottom. For German, VI and WI show a clear global peak in performance at $d = 50$, while for other languages we find several local peaks. In German, highest performance is obtained by VI, closely followed by WI, while in Swedish OP is best. In all languages performance of OP is very consistent across different $d > 25$, and it is the most robust model in high $d$. WI also shows robust performance in high $d$, but is mostly outperformed by OP. In all languages, we see a steep drop-off in performance of VI with higher $d$.

\paragraph{What is each model's optimal dimensionality?} For VI and WI the optimal $d$ in our models is much lower than the common choice of 200--300 \cite{Hamilton2016b,dubossarskyetal19,Schlechtwegetal19,Shoemark2019}.\footnote{Find an overview in Table \ref{tab:scoretask2} Post-Evaluation} In Swedish, which has the largest training corpus, optimal $d$s are higher, suggesting a possible correspondence between corpus size and optimal $d$. 

\begin{figure}
    \begin{subfigure}{0.33\textwidth}
        \includegraphics[width=\linewidth]{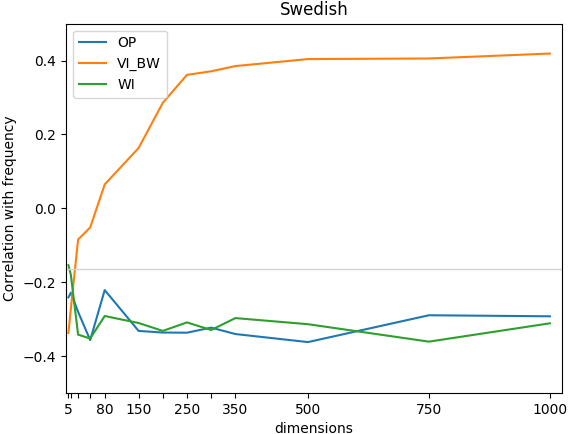}
        \caption*{2 epochs}
        \label{fig:SWEfreqcorrep2}
    \end{subfigure}
    \begin{subfigure}{0.33\textwidth}
        \includegraphics[width=\linewidth]{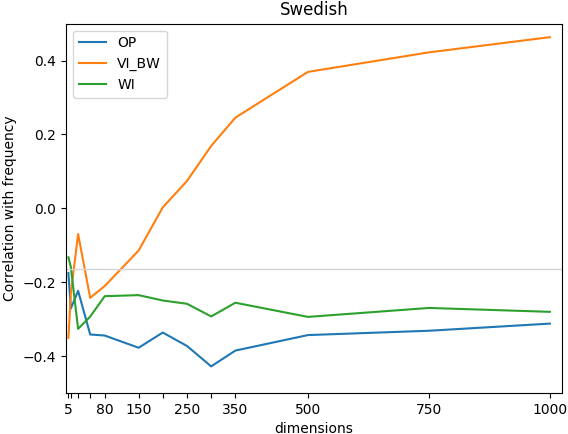}
        \caption*{5 epochs}
        \label{fig:SWEfreqcorrep5}
    \end{subfigure}
    \begin{subfigure}{0.33\textwidth}
        \includegraphics[width=\linewidth]{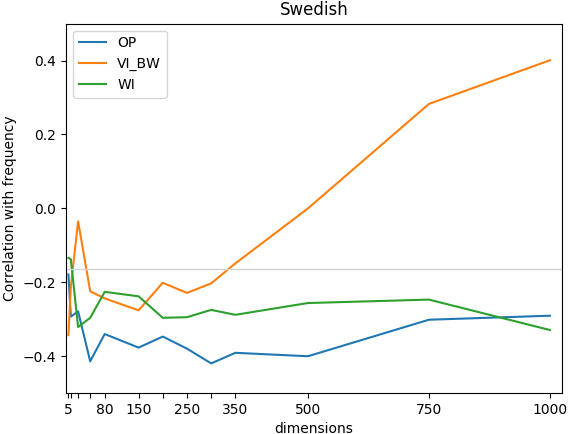}
        \caption*{10 epochs}
        \label{fig:SWEfreqcorrep10}
    \end{subfigure}
    \begin{subfigure}{0.33\textwidth}
        \includegraphics[width=\linewidth]{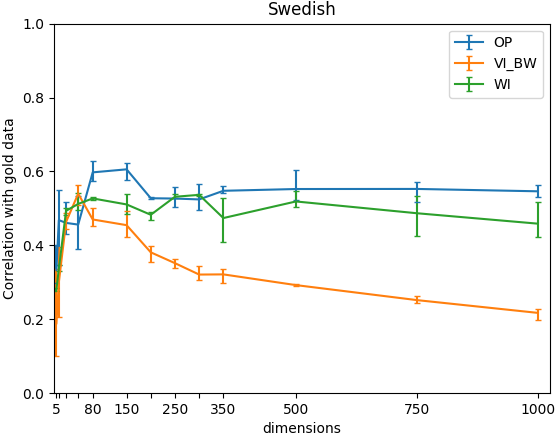}
        \caption*{2 epochs}
        \label{fig:SWEep2}
    \end{subfigure}
    \begin{subfigure}{0.33\textwidth}
        \includegraphics[width=\linewidth]{plots/task2/SWEbaselineBW.png}
        \caption*{5 epochs}
        \label{fig:SWEep5}
    \end{subfigure}
    \begin{subfigure}{0.33\textwidth}
        \includegraphics[width=\linewidth]{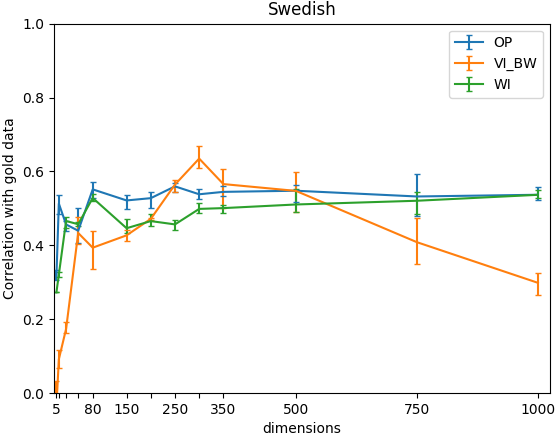}
        \caption*{10 epochs}
        \label{fig:SWEep10}
    \end{subfigure}
    \vspace{+1mm}
    \caption{\textbf{Top} figures show correlations between CD and frequency in Swedish corpus$_2$ across $d$, with increasing numbers of epochs. Gray lines indicate true correlation in gold data. \textbf{Bottom} figures show performance on Swedish data for Subtask 2.}\label{fig:freqcorr}
\end{figure}

Across all corpora, OP tends to have higher optimal values than VI and WI. This may be because orthogonal alignment works better in high dimensions, as there are more degrees of freedom to rotate the vectors. In Swedish, all methods show two local maxima instead of a global one. This behavior becomes more pronounced with increasing numbers of training epochs (see Figure \ref{fig:freqcorr}). This could be explained by the two Swedish corpora having different optimal $d$s (due to their different sizes and homogeneity).

\paragraph{Why does VI's performance drop in high $d$?} We tested several hypotheses to explain the drop. We first speculated that the embeddings for corpus$_2$ generally drift away from their initial state, as seen in \cite{Kim14} where even the word with the least amount of measured semantic change had cosine distance scores of almost 0.1. This drift could be repaired by post-hoc alignment. However, additional OP alignment did not eliminate the drop. We then tested the hypothesis that the number of training updates influences cosine distances \cite{Schlechtwegetal19}. Hence, we calculated the correlation between the predicted ranking of target words (according to cosine distance) and their frequency (reflecting the number of training updates) in the second corpus, and compared this correlation across $d$ (see Figure \ref{fig:freqcorr}, top). 
There is a clear frequency bias for VI becoming stronger with increasing $d$. Thus, the predicted cosine distances do not reflect LSC but rather frequency, leading to poor performance. The bias correlates negatively with performance as comparing the top to the bottom figures. Interestingly, the number of training epochs also has a strong effect on this bias: An increasing $e$ reduces the bias for low $d$ and consequently drastically improves performance (see Figure \ref{fig:freqcorr}, cf. top and bottom). We were not able to find an explanation for the cause of this behaviour.

The extend of the frequency bias is determined by several parameters, the main one being word frequency in the second corpus. We experimented with modified corpora wherein the frequency of all target words was fixed to 200, and less frequent target words were ignored. This completely removed the frequency bias. However, if frequency differences amongst target words exist, as is the case with most data sets, the noise induced by those differences may be exaggerated by dimensionality or reduced by number of training epochs. Understanding and explaining the frequency bias is outside of the scope of this work, but will be part of future work.

\vspace{+2mm}
\section{Conclusion}

Our shared task system investigated Vector Initialization (VI) alignment in a commonly used LSC detection model based on Skip-Gram with Negative Sampling, while focusing on the role of vector-space dimensionality. Our results suggest that LSC detection models integrating vector-space alignment should pay more attention to model-specific characteristics and the dimensionality parameter in particular. 
Current state-of-the-art models are often dominated by applying OP and WI to alignment, as a wide variety of reasonable parameters yield good results, whereas VI is very susceptible to parameters like training order, dimensionality and epochs. However we demonstrate that VI is able to outperform OP and WI alignment if tuned properly. Due to time limitations we could not fully explore the effects of epochs on VI, which have proven to play a significant role for dimensionality-dependent performance. Future work will include a closer look at the connection between dimensionality, frequency noise and training epochs.

\section*{Acknowledgments}
Dominik Schlechtweg was supported by the Konrad Adenauer Foundation and the CRETA center funded by the German Ministry for Education and Research (BMBF) during the conduct of this research. We would like to thank our reviewers for their insightful feedback.

\newpage
\bibliographystyle{coling}
\bibliography{biblio}

\newpage
\appendix
\captionsetup[figure]{skip=10pt}

\section{Details on Corpora}
\label{sec:corpora}

Table \ref{tab:sizes} shows basic corpus statistics.
\begin{table}[h]
\centering
\small
\begin{tabular}{lllllllll}
\hline
            & $t_1$ & $t_2$  & tokens$_1$ & tokens$_2$ & types$_1$ & types$_2$ & TTR$_1$ & TTR$_2$\\
\hline
\textbf{English}  & CCOHA 1810--1860  &  CCOHA 1960--2010 & 6.5M & 6.7M & 87k & 150k & 13.38 & 22.38 \\
\textbf{German} & DTA 1800--1899  & BZ+ND 1946--1990 & 70.2M & 72.3M & 1.0M & 2.3M  & 14.25 & 31.81\\
\textbf{Latin} & LatinISE -200--0 & LatinISE 0--2000 & 1.7M & 9.4M & 65k & 253k & 38.24 & 26.91 \\
\textbf{Swedish}   & Kubhist 1790--1830 & Kubhist 1895--1903  & 71.0M & 110.0M & 3.4M & 1.9M & 47.88 & 17.27\\
\end{tabular}
\vspace{+2mm}
\caption{Corpus statistics. The SemEval corpora are samples from CCOHA \cite{Davies:2012,Alatrashetal20}, DTA \cite{dta2017}, BZ \cite{BZ2018}, ND \cite{ND2018}, LatinISE \cite{mcgillivray-kilgarriff} and KubHist \cite{KubHist}.\\TTR = Type-Token ratio (number of types / number of tokes * 1000).}\label{tab:sizes}
\end{table}

\section{Pre-Processing Details}
\label{sec:process}

Table \ref{tab:threshold} shows our thresholds for infrequent words that we removed from the corpora.
\begin{table*}[htp]
	\center
\small
	\begin{tabular}{ c | c c c c | c c c c }
		\hline
	\multirow{2}{*}{} &\multicolumn{4}{c |}{\textbf{corpus$_1$}} & \multicolumn{4}{c}{\textbf{corpus$_2$}}  \\	
	 & English & German & Latin & Swedish & English & German & Latin & Swedish \\\hline
	threshold & 4 & 39 & 1 & 42 & 4 & 39 & 6 & 65\\
		\hline
	\end{tabular}
    \vspace{+2mm}
	\caption{Corpus-specific frequency thresholds.}
	\label{tab:threshold}
\end{table*}

\section{Influence of training order}
\label{sec:order}

\begin{figure}[H]
    \begin{subfigure}{0.5\textwidth}
        \includegraphics[width=\linewidth]{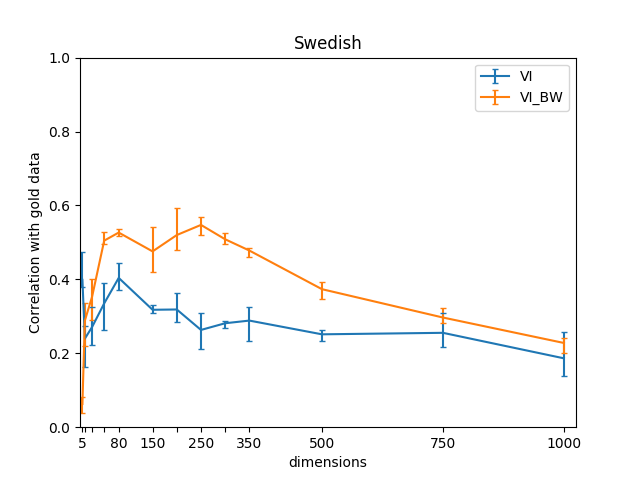}
                \label{fig:SWEVIBW}
    \end{subfigure}
    \begin{subfigure}{0.5\textwidth}
        \includegraphics[width=\linewidth]{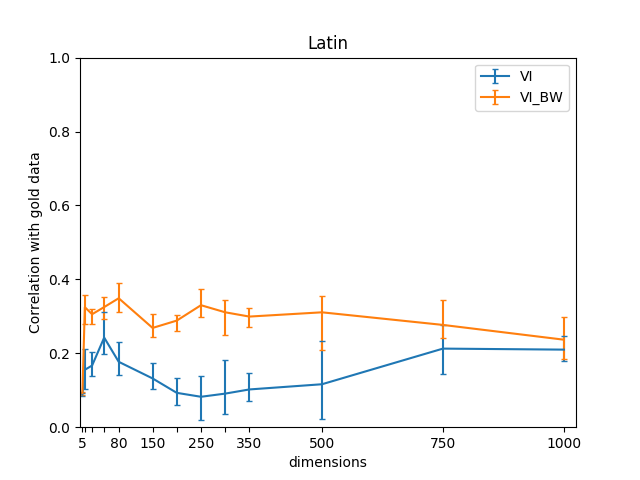}
                 \label{fig:LTNVIBW}
    \end{subfigure}
    \caption{Comparing performances on Swedish and Latin when changing the training order.\\VI: normal order, VI\_BW: backwards order}
    \label{fig:backwards}
    \vspace{+5mm}
\end{figure}

We switched the training order for Swedish and Latin due to their size differences across corpora. The switch led to noticeable increases in performance for both languages, see Figure \ref{fig:backwards}.

\end{document}